\documentclass[arxiv]{article}

\usepackage{soul}
\usepackage{url}
\usepackage{float}
\usepackage[utf8]{inputenc}
\usepackage{wrapfig}
\usepackage{graphicx}
\usepackage{epstopdf}
\usepackage{amsmath}
\usepackage{amssymb}
\usepackage{booktabs}
\usepackage{algorithm}
\usepackage[noend] {algorithmic}
\usepackage{enumitem}
\usepackage{booktabs} 
\usepackage{color,soul,xspace}
\usepackage{hyperref}

\usepackage{comment}
\usepackage{epsfig}
\usepackage{subfig}
\usepackage{mathrsfs,mdwlist,enumitem}
\usepackage{stmaryrd}

\newcommand{\name}{\textsc{BroGNet}\xspace}
\newcommand{\bfgn}{\textsc{BFgn}\xspace}
\newcommand{\bdgnn}{\textsc{BDGnn}\xspace}
\newcommand{\bnn}{\textsc{Bnn}\xspace}

\newcommand{\gnn}{\textsc{Gnn}\xspace}

\newcommand{\MLP}{\texttt{MLP}\xspace}
\newcommand{\sqp}{\texttt{squareplus}\xspace}

\newcommand{\CV}{\mathcal{V}\xspace}
\newcommand{\CE}{\mathcal{E}\xspace}

\newcommand{\cW}{\mathbf{W}\xspace}

\newcommand{\ch}{\mathbf{h}\xspace}

\newcommand{\cz}{\mathbf{z}\xspace}

\setlist{nolistsep,leftmargin=*}

\newtheorem{thm}{\textbf{Theorem}}

\newcommand{\rev}[1]{\textcolor{black}{#1}}

\usepackage[utf8]{inputenc} 
\usepackage[T1]{fontenc}    
\usepackage{hyperref}       
\usepackage{url}            
\usepackage{booktabs}       
\usepackage{amsfonts}       
\usepackage{nicefrac}       
\usepackage{microtype}      
\usepackage{xcolor}         

\title{Graph Neural Stochastic Differential Equations for Learning Brownian Dynamics}

\author{Suresh Bishnoi, Jayadeva, Sayan Ranu, N. M. Anoop Krishnan\thanks{SB: School of Interdisciplinary Research, SR: Department of Computer Science, NMAK and RB: Department of Civil Engineering, J: Department of Electrical Engineering, SR and NMAK: Yardi School of Artificial Intelligence (joint appointment).}\\
Indian Institute of Technology Delhi, Hauz Khas, New Delhi, India 110016\\
\texttt{\{srz208500,jayadeva,sayanranu,krishnan\}@iitd.ac.in}
}

\begin{document}

\maketitle

\begin{abstract}
Neural networks (NNs) that exploit strong inductive biases based on physical laws and symmetries have shown remarkable success in learning the dynamics of physical systems directly from their trajectory. However, these works focus only on the systems that follow deterministic dynamics, for instance, Newtonian or Hamiltonian dynamics. Here, we propose a framework, namely \textit{Brownian graph neural networks} (\name), combining \textit{stochastic differential equations (SDEs)} and \gnn{s} to learn Brownian dynamics directly from the trajectory. We theoretically show that \name conserves the linear momentum of the system, which in turn, provides superior performance on learning dynamics as revealed empirically. We demonstrate this approach on several systems, namely, linear spring, linear spring  with binary particle types, and non-linear spring systems, all following Brownian dynamics at finite temperatures.  We show that \name significantly outperforms proposed baselines across all the benchmarked Brownian systems. In addition, we demonstrate zero-shot generalizability of \name to simulate unseen system sizes that are two orders of magnitude larger and to different temperatures than those used during training. 
 Altogether, our study contributes to advancing the understanding of the intricate dynamics of Brownian motion and demonstrates the effectiveness of graph neural networks in modeling such complex systems.
\end{abstract}

\section{Introduction and Related Works}
Learning the dynamics of physical systems directly from their trajectory is an active area of research due to their potential applications in materials modeling~\cite{park2021accurate}, drug discovery~\cite{vamathevan2019applications}, motion planning~\cite{ni2022ntfields}, robotics~\cite{sanchez2019hamiltonian,greydanus2019hamiltonian}, and even astrophysics~\cite{sanchez2020learning}. Recent works demonstrated that physics-based inductive biases could enable the learned models to follow conservation laws such as energy and momentum while also simplifying the learning making the model data efficient~\cite{benchmarking,lnn1}. Among these, a family of models, such as Lagrangian or Hamiltonian neural networks~\cite{lgnn,rigidbody,metahamiltonian,sanchez2019hamiltonian,greydanus2019hamiltonian} and Neural ODEs~\cite{gnode,zhong2019symplectic,gruver2021deconstructing,chen2018neural}, enforces the physics-based inductive biases in a strong sense. Here, a governing ordinary differential equation (ODE) is used along with a neural network to learn the abstract quantities, such as energy or force, directly from the trajectory of the system. These models have shown remarkable success in learning the dynamics of a variety of systems in an efficient fashion, \textit{viz.}, particle-based systems~\cite{gnode,benchmarking}, atomistic dynamics~\cite{park2021accurate,huang2021equivariant}, physical systems~\cite{lgnn,sanchez2019hamiltonian,greydanus2019hamiltonian}, and articulated systems~\cite{rigidbody}.

Despite their success, these works focus on purely deterministic systems where the dynamics are governed by an ODE~\cite{karniadakis2021physics,benchmarking,zhong2021benchmarking}. An alternative approach to model physical systems is to formulate the governing equation as a \textit{stochastic} differential equation (SDE), for instance, \textit{Brownian} dynamics (BD)---widely used to study the dynamics of particles in a fluid or solvent~\cite{van1982algorithms}. Brownian dynamics simulations have proven to be highly valuable in numerous scientific disciplines. In physics, they have been used to study the diffusion of particles~\cite{chung1999permeation}, the self-assembly of colloidal systems~\cite{noguchi2001self}, and the behavior of polymers in solution~\cite{helfand1980brownian}. In chemistry, Brownian dynamics simulations have shed light on reaction kinetics~\cite{northrup1992kinetics}, molecular diffusion~\cite{madura1995electrostatics}, and the behavior of macromolecules~\cite{cates1985brownian}. In biology, they have provided insights into the movement of cells~\cite{klank2018brownian}, the folding of proteins~\cite{karplus1994protein}, and the dynamics of biomolecular complexes~\cite{gabdoulline2001protein}. Despite their importance in such wide domains, while there have been approaches to inferring the nature of Brownian dynamics from observations in a statistical sense~\cite{gnesotto2020learning,genkin2021learning}, to the best of the authors' knowledge, no attempt has been made to learn the dynamics of Brownian systems in particular, or SDEs in general, from their trajectory.

Here, we propose \name, which can learn the Brownian dynamics of a system directly from the trajectory. Exploiting a \gnn-based framework, we show that the Brownian dynamics can be learned from small systems, which can then be generalized to arbitrarily large systems. The major contributions of the work are as follows.
\begin{itemize}
    \item \textbf{Learning stochastic dynamics from trajectory:} We present a framework, \name, that learns the abstract quantities governing the dynamics of stochastic systems directly from their trajectory. While demonstrated for Brownian dynamics, the framework presented is generic and may be applied to other SDEs as well.
    \looseness=-1
    \item \textbf{Graph-based modeling for zero-shot generalizability:} We model the physical system as a graph with nodes representing particles and edges representing their interactions, which can then be used to learn BD. The inductive graph-based modeling naturally imparts the desirable properties of \textit{zero-shot generalizability} to unseen system sizes and \textit{permutation invariance.} In addition, we also demonstrate generalizability to unseen temperatures in a zero-shot fashion.
    \item \textbf{Momentum conservation:} We demonstrate that the proposed framework follows Newton's third law, thereby conserving the total linear momentum of the system. This feature, in turn, provides superior performance for the model.
    \looseness=-1
\end{itemize}

\section{Background on Brownian Dynamics}
The fundamental concept behind Brownian dynamics is the influence of thermal fluctuations on the motion of particles. In a fluid, particles are subject to random collisions with the surrounding molecules, which lead to erratic motion characterized by continuous changes in direction and speed (see the videos in the supplementary for a visual depiction). The behavior of particles undergoing Brownian motion can be described mathematically using SDEs, such as over-damped \textit{Langevin} equations~\cite{langevin}~\cite{branka1998algorithms}. These equations incorporate deterministic forces, such as external potentials or interactions between particles, as well as stochastic forces that represent the random effects of the surrounding fluid. By numerically integrating these equations, it is possible to simulate and analyze the dynamic behavior of the particles over time.

Consider a system of $N$ particles having masses $M$ and interacting with each other through an interaction potential $U(X_t)$, where $X_t$ represents the position vectors of all the particles at time $t$, that constitute a time-dependent random variable. The Langevin equation governing the dynamics of this system is given by~\cite{langevin}
\begin{equation}
\label{eq:langevin}
    M\ddot{X_t} = -\nabla U(X_t)-\zeta M \dot{X_t} + \sqrt{2 M \zeta k_B T}~\Omega_t
\end{equation}
where $-\nabla U(X_t) = F(X_t)$ is the force due to the inter-particle interactions, $\zeta$ is the damping constant, $T$ is the temperature, $k_B$ is the Boltzmann constant, and $\Omega_t$ is a delta-correlated stationary Gaussian process with zero mean that follows $\langle \Omega_t \rangle = 0$ and $\langle \Omega_t.\Omega_{t'} \rangle = \delta(t-t')$, where $\delta$ is the standard Dirac-delta function. In the over-damped limit of Langevin dynamics, where no acceleration exists, the equation reduces to Brownian dynamics as~\cite{branka1998algorithms}
\begin{equation}
\label{eq:langevinsimple}
\gamma \dot{X_t} = -\nabla U(X_t) + \sqrt{2 \gamma k_B T}~\Omega_t
\end{equation}
where $\gamma = M \zeta$. Thus, the evolution of a system can be obtained by numerically integrating the SDE given by Eq.~\ref{eq:langevinsimple}. Note that the first term in the RHS of Eq.~\ref{eq:langevinsimple} is a deterministic force term, and the randomness in the dynamics comes from the second term which is stochastic in nature.

\section{\name: A Graph Neural SDE Approach}
\begin{figure}
\centering
\includegraphics[width=\columnwidth]{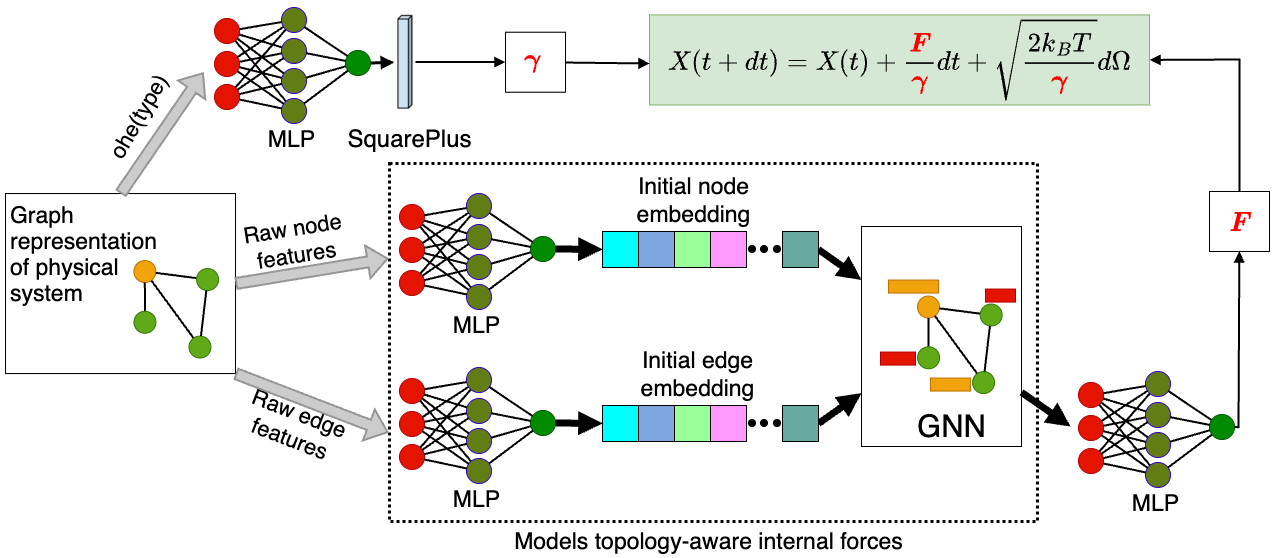}
\caption{\name architecture.}
\label{fig:graph_architecture}
\end{figure}
To learn the trajectories of a multi-particle system governed by Brownian dynamics, we design a neural network called \name. 
 \name is essentially a Graph Neural SDE---inspired by Graph Neural ODE, which parametrizes an ODE using a \gnn{}. Thus, \name parameterizes the stochastic dynamics $B(X_t)$ using a neural network and learns the approximate function $\hat{B}(X_t)$ by minimizing the loss between the \textit{distribution} of ground truth trajectories with that of the predicted trajectories. However, in contrast to classical dynamics, the loss is applied over a distribution of positions rather than the exact deterministic trajectory due to the stochastic nature of the dynamics. We discuss the exact loss function in detail later. 
 
 The learning process is, therefore, decomposed into two factors: the \textit{standard deviation} associated with the ground-truth stochastic process corresponding to each particle and their \textit{expected positions}. In \name, the standard deviation is learned through an \MLP. On the other hand, the expected positions are learned through a Graph Neural SDE. The choice of using a \gnn to model interaction dynamics is motivated by two key observations. First, a \gnn allows us to be inductive, where the number of model parameters is independent of the number of particles in the system. Hence, it enables zero-shot generalizability to systems of arbitrary unseen sizes. In addition, a \gnn allows \textit{permutation-invariant} modeling of the system. Below, we discuss the details.

\subsection{The \gnn Architecture}
\label{sec:gnn}
\vspace{-0.10in}

\name models the interaction dynamics among participating particles as a graph. The graph topology is used to learn the approximate dynamics $\hat{B}$ by minimizing the distance between distributions representing the true and predicted positions. Note that henceforth all the variables with a hat ( $\hat{}$ ) correspond to the approximate variable learned by the machine learning model.

{\bf Graph structure.} We represent an $N$-particle system as an directed graph $\mathcal{G}=\{\mathcal{V,E}\}$. $\CV$ represents the set of nodes, where each node represents a particle. Similarly, $\CE$ represents interactions or connections among particles in the form of edges. For example, in a ball-spring system or colloidal gels, the balls or colloid particles correspond to nodes, and the springs or the interactions between the colloidal particles correspond to edges. Note that in the case of systems such as springs, where explicit connections are pre-defined, the graph structure remains static. However, in the case of systems like colloidal gels, the edges are defined based on a neighborhood cut-off, and hence the graph structure can be dynamic, with the edges between particles changing as a function of the particle configuration. As we will see in the next paragraph, a directed graph is required only because the edge weights we use are directional. 

{\bf Input features.} Each node corresponds to a particle $i$ and is defined by its intrinsic features, such as the particle's type ($\tau_i$), position ($X_{i,t}$), and velocity ($\dot X_{i,t}$) at time $t$. The position and velocities are 3-dimensional tuples corresponding to the $x$, $y$ and $z$ dimensions. The particle type distinguishes between different characteristics, such as varying masses or friction coefficient for balls or particles. For brevity, $X_{i,t}$ is referred to as $X_{i}$ henceforth unless specified otherwise. On the other hand, each edge represents the connection between nodes and is represented by the 3-dimensional edge feature $w_{ij}=(X_i-X_j)$, which signifies the relative displacement of the connected nodes along each coordinate. Note that the edge weights are asymmetric due to the vectorial nature of the input.

\textbf{Summary of neural architecture.} Fig.~\ref{fig:graph_architecture} provides a pictorial description of \name. In the pre-processing layer, we utilize Multilayer Perceptrons (MLPs) to create dense vector representations for each node $v_i$ and edge $e_{ij}$. In systems where internal forces play a significant role in governing dynamics, the structure's topology holds crucial importance. To capture this dependency, we employ a deep, message-passing Graph Neural Network (\gnn). The final representation $\cz_i=\ch_i^L$ is generated by the \gnn for each node, capturing relevant information. These representations are subsequently passed through another \MLP to predict the force of each node. We next detail these individual steps.

\textbf{Pre-processing.} In the pre-processing layer, a dense vector representation is constructed for each node $v_i$ and edge $e_{ij}$ using an MLP denoted as $\texttt{MLP}_{em}$. The construction process can be described as
\begin{alignat}{2}
\ch^0_i &= \sqp(\texttt{MLP}_{em}(\texttt{one-hot}(\tau_i))) \label{eq:one-hot}\\
\ch^0_{ij} &= \sqp(\texttt{MLP}_{em}(w_{ij}))
\end{alignat}

Here, $\sqp$ represents an activation function. It is important to note that the node and edge embedding functions, parametrized by $\texttt{MLP}_{em}$, utilize separate weights. For the sake of brevity, we refer to them simply as $\texttt{MLP}_{em}$. Note that the node representations do not explicitly encode their ground-truth positions. Only the distances between nodes are encoded in the edge representations. This allows us to ensure \textit{translational invariance.}

\textbf{Force prediction.} We utilize multiple layers of message-passing to facilitate communication between nodes and edges. In the $\ell^{th}$ layer, the node embeddings are updated as follows:
\begin{equation}
\ch_i^{\ell+1} = \texttt{squareplus} \left( \cW_{\CV}^{\ell}\cdot\left( \ch_i^{\ell}\bigparallel\sum_{j \in \mathcal{N}^{in}_i} \ch_{ji}^{\ell} \bigparallel \sum_{j \in \mathcal{N}^{out}_i}  \ch_{ij}^{\ell}\right)\right)
\end{equation}
Here, $\mathcal{N}^{in}_i=\{v_j\in\CV\mid e_{ji}\in\CE \}$ and $\mathcal{N}^{out}_i=\{v_j\in\CV\mid e_{ij}\in\CE \}$ represents the incoming and outgoing neighbors of node $v_i$ respectively. The term $\cW_{\CV}^{\ell}$ signifies a layer-specific learnable weight matrix. More simply, we sum-pool the edge embeddings over both incoming and outgoing neighbors, concatenate them along with the target node's own embedding and then pass it through one linear transformation followed by non-linearity. The embedding of an edge $e_{ij}$ on node $v_i$ in the $\ell^{th}$ layer, denoted as $\ch_{ij}^{\ell}$, is computed as:
\begin{equation}
\ch_{ij}^{\ell+1} = \texttt{squareplus} \left( \cW_{\CE}^{\ell}\cdot \left(\ch_{ij}^{\ell} \bigparallel \ch_i^{\ell} \bigparallel \ch_{j}^{\ell} \right)\right)
\end{equation}
Similar to $\cW_{\CV}^{\ell}$, $\cW_{\CE}^{\ell}$ is a layer-specific learnable weight matrix that applies to the edge set. The message-passing process spans $L$ layers, with $L$ being a hyper-parameter. In the $L^{th}$ layer, the final node and edge representations are denoted as $\cz_i=\ch_i^L$ and $\cz_{ij}=\ch_{ij}^L$, respectively.

Finally, the pair-wise interaction force $\hat{F}_{ij}$ from particle $i$ to $j$ is predicted as:
\begin{equation}
\hat{F}_{ij}=\texttt{squareplus}(\texttt{MLP}_{\CV}(\cz_{ij}))
\end{equation}
where, $\texttt{MLP}_{\CV}$ denotes a Multilayer Perceptron with a squareplus activation function. 

The force on a node is defined as the summation of all forces from incoming edges and the \textit{reactive force} from its outgoing edges. Mathematically,
\begin{equation}
\hat{F}_i = \sum_{j \in \mathcal{N}^{in}_i} \hat{F}_{ji} + \sum_{j \in \mathcal{N}^{out}_i} -\hat{F}_{ij}
\end{equation}

Here, the second term accounts for the reactive force from each outgoing edge, which is essentially the negated version of the force particle $i$ imparts on particle $j$. This inductive bias ensures Newton's third law that every action has an equal and opposite reaction. More importantly, this inductive bias implicitly ensures that the net force on the system is zero, thereby strictly ensuring momentum conservation. Formally:
\begin{thm}[Momentum Conservation] \textit{In the absence of an external field, \name exactly conserves the linear momentum of the system.}
\label{thm:mom_cons}
\end{thm}
Proof of the theorem is provided in Appendix~\ref{app:proof}. Further, as demonstrated later the empirical results comparing the performance of \name with a version that does not employ the momentum conservation, namely, Brownian dynamics \gnn{} (\bdgnn), suggests that the momentum conservation indeed results in better performance of the model.

\textbf{Learning the Brownian term.} While the deterministic force term (first term in the RHS of Eq.~\ref{eq:langevinsimple}) in $\hat{B}$ is learned by the \gnn{} as outlined above, the stochastic term (second term in Eq.~\ref{eq:langevinsimple}) is independent of the system topology and depends only the particle type and its attributes such as radius or friction coefficient. To learn this term, we represent the node type as a one-hot encoding, which is passed through an \MLP to obtain the $\hat{\gamma}_i$ associated with each particle (see Fig.~\ref{fig:graph_architecture}). This  $\hat{\gamma}_i$, along with force obtained from the \gnn{}, is substituted in the governing equation (Eq.~\ref{eq:langevinsimple}) and learned in an end-to-end fashion. It is worth noting that the standard deviation of the stochastic process is given by $\sigma_i = \sqrt{2\gamma_i k_B T}$. Thus, all other terms other than $\gamma_i$ in the Eq.~\ref{eq:langevinsimple} are known apriori or are constants. 

\textbf{Trajectory prediction and training.} Based on the predicted forces and $\gamma_i$s, the positions are derived using the \textit{Euler Maruyama} integrator~\cite{branka1998algorithms}. The loss on the prediction is computed using \textit{Gaussian Negative Log-likelihood} loss, which is then back-propagated to train the model. Specifically, let $X_{i,t}$ and $\hat{X}_{i,t}$ denote the ground truth and predicted positions of particle $i$ at time $t$, respectively. The loss function is defined as follows.
\begin{equation}
    \label{eq:lossfunction}
  \mathcal{L}= \frac{1}{N}\left(\sum_{i=1}^N \sum_{t=2}^T \log \max\{\hat{\sigma}_i^2,\epsilon\}+\lambda\frac{(X_{i,t}-\hat{X}_{i,t})^2}{\max\{\hat{\sigma}_i^2,\epsilon\}}\right)
\end{equation}
 Here, $\epsilon$ is a small constant, and $\lambda$ controls the relative weightage of the two terms. $\hat{\sigma}_i$ is the standard deviation associated with the position of each of the particles. The overall objective of the loss is to minimize the variance-normalized deviation from the ground truth (second term), while also ensuring a regularizing term to keep the variance low. We use Gaussian NLL loss since the stochasticity in Brownian dynamics is a Gaussian as well (recall~Eq.~\ref{eq:langevinsimple}). 
 
 Note that the positions, both predicted and ground-truth, are computed directly from the predicted and ground-truth forces, respectively, using \textit{Euler Maruyama} integrator as follows.
 \begin{equation}
 X_i(t+\Delta t) = X_i(t) + F_i/\gamma_i \Delta t + \sqrt{\frac{2k_BT}{\gamma_i}} \Delta \Omega_t
 \label{eq:em_int}
 \end{equation}
where $\Delta \Omega_t$ is a random number sampled from a standard Normal distribution.
It should be noted that the training approach presented here may lead to learning the dynamics as dictated by the \textit{Euler Maruyama} integrator. Thus, the learned dynamics may not represent the ``true'' dynamics of the system, but one that is optimal for the \textit{Euler Maruyama} integrator\footnote{\rev{See App.~\ref{app:em_int} for details on \textit{Euler Maruyama} integrator and other integrators for SDE.}}.


\section{Experiments}
\label{sec:experiments}
In this section, we benchmark the ability of \name to learn Brownian dynamics directly from the trajectory and establish:
\begin{itemize}
\item {\bf Accuracy:} \name accurately models Brownian dynamics and outperforms baseline models.
\item {\bf Zero-shot generalizability:} The inductive architecture of \name allows it to accurately generalize to systems of unseen sizes and temperatures.
\end{itemize}
All the codes used in the present work will be made available upon the acceptance of the work.

\subsection{Experimental setup}
$\bullet$ \textbf{Simulation environment.} All the training and forward simulations are carried out in the JAX environment~\cite{schoenholz2020jax}. The graph architecture is implemented using the jraph package~\cite{jraph2020github}.\\ 
\textbf{Software packages:} numpy-1.20.3, jax-0.2.24, jax-md-0.1.20, jaxlib-0.1.73, jraph-0.0.1.dev0 Hardware: Memory: 16GiB System memory, Processor: Intel(R) Core(TM) i7-10750H CPU @ 2.60GHz.\\
$\bullet$ \textbf{Baselines:} To the best of the authors' knowledge, there are no prior works on learning Brownian dynamics directly from the trajectory. Thus, in order to compare the performance of \name, we propose four different baselines.\\ 
(1) \textbf{Neural Network (NN):} Here, we use a feedforward \MLP that takes the $X_i$ and $\dot{X}_i$ concatenated into a single vector at time $t$ as the input and predicts the positions and velocities at $t + \Delta t$ as the output.\\ 
(2) \textbf{Brownian NN (\bnn):} NN in (1) does not have any physics-based inductive biases. To address this, we propose \bnn, a Neural SDE that parameterizes the Brownian dynamics using a feed-forward \MLP. Specifically, \bnn takes $X_t$ of all the particles concatenated into a vector as input and predicts the $\hat{F}$ and $\hat{\gamma}$ as output employing a fully connected \MLP. These outputs are then used in Eq.~\ref{eq:langevinsimple} and Eq.~\ref{eq:em_int} to predict the updated positions and velocities. Thus, \bnn employs the inductive-bias in terms of the Brownian equation of motion.\\ 
(3) \textbf{Brownian full graph network (\bfgn):} The models in (1) and (2) do not exploit the topology of the system and are not permutation invariant. To this extent, we employ \bfgn a graph-based baseline that takes the position and velocities as input and outputs the forces and $\gamma$ for each node. This output is then substituted in Eq.~\ref{eq:langevinsimple} to obtain the updated position. Thus, \bfgn is a graph-based version of \bnn. Note that \bfgn employs the full graph network architecture~\cite{cranmer2020discovering,sanchez2020learning}, a versatile architecture that has been demonstrated for a variety of particle-based systems. It is worth noting that \bfgn is not translational invariant due to the explicit use of positions as node inputs.\\ 
(4) \textbf{Brownian Dynamics \gnn (\bdgnn):} Finally, we also present a physics-informed baseline, \bdgnn. \bdgnn has all the inductive biases of \name except Newton's third law. Specifically, the force on each particle $i$ is computed as
\begin{equation}
    F_{i}=\texttt{squareplus}(\texttt{MLP}_{\CV}(\cz_{i}))
\end{equation}
Note that \bdgnn has the same architecture as \name and hence allows us to evaluate the role of momentum conservation as an inductive bias in the \name.

\textbf{$\bullet$ Datasets and systems:} To compare the performance \name, we selected three different systems, namely, linear spring, linear spring  with binary particle types, and non-linear spring systems, all following Brownian dynamics at finite temperatures. The first is a model system with particles interacting with each other based on a linear spring and subjected to Gaussian noise. The second system comprises of non-linear spring where the force on a spring is proportional to the fourth power of displacement. The third system corresponds to a linear spring system with two different types of particles having different $\gamma$. All the graph-based models are trained on five spring systems only, which are then evaluated on other system sizes. The details of all the systems to generate the ground truth data are provided in App.~\ref{app:systems}. Further, the detailed data-generation procedure is given in App.~\ref{app:imple}.

\textbf{$\bullet$ Evaluation Metric:} To quantify performance, we use the following metrics:
\begin{itemize}
\item {\bf Position error: }The position error is the normalized Euclidean distance between the ground truth and predicted positions. Mathematically, let $(x,y,z)$ and $\hat{x},\hat{y},\hat{z}$ be the ground truth and predicted positions, respectively, of a given particle. Furthermore, let $\sigma_{x/y/z}$ be the ground-truth standard deviation of the positions of the particle. The error is, therefore:
\begin{equation}
\label{eq:pe}
PE=\left(\sum_{c\in\{x,y,z\}}\left(\frac{c-\hat{c}}{\sigma_c}\right)^2\right)^{1/2}
\end{equation}
\item {\bf Brownian error: }The Brownian error quantifies the difference between the standard deviation estimated by the neural network with that of the ground truth using the RMSE metric.
\item {\bf Trajectory roll-out error: }The above two metrics individually look at the errors in positions and the standard deviation. In this metric, we holistically compare the distance between the distribution of trajectories in the ground truth against the ones predicted by \name. Specifically, given a Brownian system, we simulate $100$ ground-truth trajectories across $100$ time steps over each particle in the system. Similarly, we predict the trajectories using \name. We next compute the average distance between the ground-truth distribution and the predicted one using KL-divergence across all particles. KL-divergence is defined as follows:
\begin{equation}
\label{eq:kl}
D_{\text{KL}}(\hat{\mathcal{X}}\| \mathcal{X}) = \sum_{x \in \mathcal{\hat{X}}} \hat{\mathcal{X}}(x) \log \left( \frac{\hat{\mathcal{X}}(x)}{\mathcal{X}(x)} \right)
\end{equation}
Here, $\mathcal{X}$ and $\hat{\mathcal{X}}$ represent the distributions over the ground truth and predicted trajectories, respectively. Both these distributions are univariate normal distributions since the stochastic component in Brownian dynamics is normally distributed (Eq.~\ref{eq:langevinsimple}). Hence, Eq.~\ref{eq:kl} reduces to the following:
\begin{align}
D_{\text{KL}}(\hat{\mathcal{X}}\| \mathcal{X}) &= \log\frac{\sigma_1}{\sigma_0}+\frac{\sigma_0^2+(\mu_o-\mu_1)^2}{2\sigma_1^2}-\frac{1}{2}
\label{eq:D_KL}
\end{align}
Here, $\sigma_0$ and $\mu_0$ are the standard-deviation and mean of distribution $\hat{\mathcal{X}}$ and similarly $\sigma_1$ and $\mu_1$ denote the standard deviation and mean of $\hat{\mathcal{X}}$.
\end{itemize}

\textbf{$\bullet$ Model architecture and training setup:} 
We use 10000 data points generated from 100 trajectories with random initial conditions to train all the models. Detailed training procedures and hyper-parameters are provided in App.~\ref{app:hyper}. For all versions of the graph architectures, MLPs are two layers deep. All models were trained till the decrease in loss saturates to less than 0.001 over 100 epochs. The model performance is evaluated on a forward trajectory, a task it was not explicitly trained for, of $0.1$s. To compute the evaluation metrics, all the results are obtained over trajectories generated from 100 different initial conditions. Further, to evaluate the zero-shot generalizability, the models trained on 5 spring systems are evaluated on 50, and 500 spring systems. Further, the models trained at 1 unit temperature are evaluated at 10 and 100 unit temperatures.

\subsection{Comparison with baselines}
\begin{figure}
\centering
\includegraphics[width=\columnwidth]{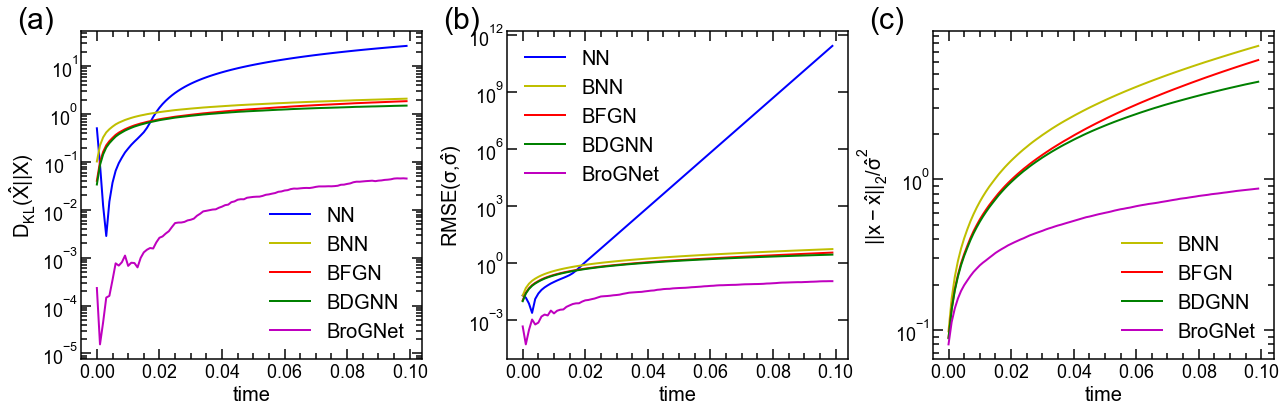}
\caption{Comparison of \name on a linear 5-spring system with baselines using (a) trajectory roll-out error, (b) Brownian error, and (c) position error. The plots are obtained based on 100 forward simulations by the trained model with random initial conditions.}
\label{fig:fig_spring_baselines}
\end{figure}
Fig.~\ref{fig:fig_spring_baselines} shows the performance of all the models on linear spring systems with 5 particles. We show that \name significantly outperforms all other models in terms of position error, Brownian error, and trajectory roll-out error. Specifically, we observe that the \name outperforms \bdgnn, which differs from the \name only in terms of momentum conservation. This suggests that the momentum conservation bias added \name significantly enhances the performance of the model. Interestingly, we also observe that the performance of \bnn, \bfgn, and \bdgnn are comparable, although \bdgnn seem to perform slightly better. This suggests that the inability of the \bfgn to ensure translational invariance has no significant impact on the model performance.

\begin{figure}
\centering
\includegraphics[width=\columnwidth]{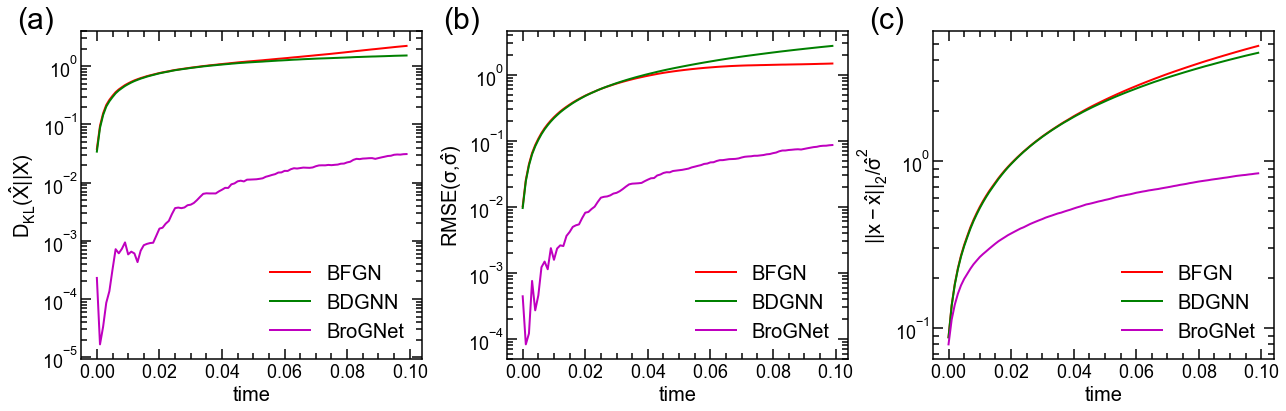}
\caption{Comparison of \name on a non-linear 5-spring system with baselines, \bfgn and \bdgnn using (a) trajectory roll-out error, (b) Brownian error, and (c) position error. The plots are obtained based on 100 forward simulations by the trained model with random initial conditions.}
\label{fig:fig_non_spring_baselines}
\end{figure}

\begin{figure}
\centering
\includegraphics[width=\columnwidth]{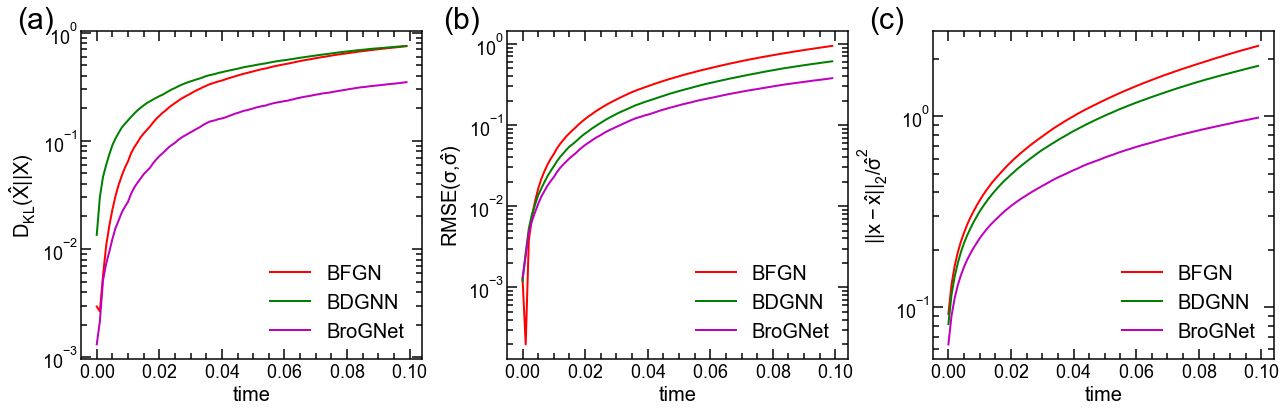}
\caption{Comparison of \name on a binary linear 5-spring system with \bfgn and \bdgnn as a baseline using (a) trajectory roll-out error, (b) Brownian error, and (c) position error. The plots are obtained based on 100 forward simulations by the trained model with random initial conditions.}
\label{fig:fig_binary_spring_baselines}
\end{figure}

Figure~\ref{fig:fig_non_spring_baselines} shows the performance of \name on non-linear spring, and Figure~\ref{fig:fig_binary_spring_baselines} shows the performance on binary linear spring systems, respectively. Note that, due to their inferior performance, the baselines NN and \bnn are not included in the figure. Thus, only the baselines \bfgn and \bdgnn are included. The complete results, including these baselines, are included in the App.~\ref{app:baselines}. As in the case of the linear spring system, we observe that \name outperforms all other models in learning the Brownian dynamics for non-linear spring systems and binary linear spring systems. However, we note that the difference between the performance of \name with other baselines is lower in binary spring systems. Nevertheless, these results suggest the ability of \name to learn the dynamics of complex multiparticle systems interacting with each other based on non-linear interaction forces.

\subsection{Zero-shot generalizability}
\begin{figure}
\centering
\includegraphics[width=\columnwidth]{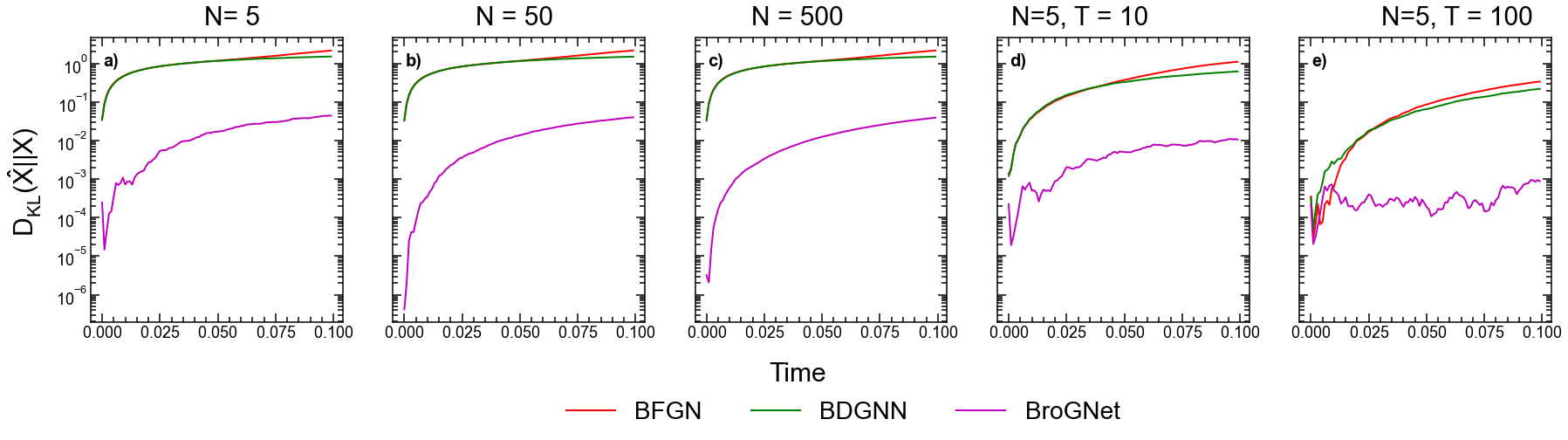}
\caption{Trajectory roll-out error of \name, \bfgn, and \bdgnn trained on N = 5 and T =1 to unseen system sizes (N = 50, and 500) and unseen temperatures (T = 10 and 100). The plots are obtained based on 100 forward
simulations by the trained model with random initial conditions.}
\label{fig:fig_generalizability}
\end{figure}

Now, we evaluate the zero-shot generalizability of the learned models to unseen system sizes and temperatures. To this extent, we use the learned models, \bfgn, \bdgnn, and \name trained on N = 5 and T = 1 units. First, these models are evaluated on system sizes of 50 and 500 springs, up to two orders of magnitude larger than those used in the training data. Figure~\ref{fig:fig_generalizability}(a-c) shows the trajectory rollout error of \name, \bfgn, and \bdgnn on 5, 50, and 500 spring systems, respectively. Interestingly, we observe all three models exhibit inductivity to larger system sizes, thanks to the \gnn{}. Further, we observe that the trajectory rollout error for all the models is comparable with the increasing system size, even when evaluated on a system size of 500. The other error metrics, namely, the position error and Brownian error are included in the App.~\ref{app:errors}. These results suggest that the \name can be trained on small datasets and generalized to arbitrarily large system sizes for performing inference on the Brownian dynamics of the system. It should be noted that both NN and \bnn, due to their architecture (feed-forward MLP), is not inductive and does not allow inference on larger system sizes.

Another important aspect of Brownian dynamics is the ability to simulate the learned systems corresponding to different thermodynamic conditions, namely, different temperatures. Figure~\ref{fig:fig_generalizability}(e,f) shows the trajectory rollout error of linear spring systems with five particles at temperatures 10 and 100 units, respectively. We again observe that \name outperforms \bfgn and \bdgnn with very low errors on the roll-out. Other error metrics, including the position error and Brownian error, as shown in App.~\ref{app:errors} also reinforce the superior performance of the \name. This suggests that both the stochastic Brownian terms and the deterministic force term learned by the \name are accurate and scalable to large system sizes and different thermodynamic conditions.


\section{Conclusion}
Altogether, in the present work, we demonstrate a graph-based neural SDE framework that can learn Brownian dynamics directly from the data. We demonstrate the approach to linear and non-linear springs and binary linear spring systems subjected to a stochastic dynamic term and following Brownian dynamics. In order to compare the performance of \name, we proposed several baselines, namely, NN, \bnn, \bfgn, and \bdgnn. Note that these baselines were proposed due to the paucity of baselines in the literature. We show that the \name architecture, with its unique momentum conservation, significantly outperforms all other baselines. Specifically, we demonstrate that momentum conservation is an important feature that can significantly enhance the performance of such models. We also demonstrate the zero-shot generalizability of the learned model to unseen system sizes and unseen temperatures. These results suggest that \name presents a robust framework to learn Brownian dynamics from small amounts of data.

\textbf{Broader Impact:} Brownian dynamic simulations play an important role in various applications including drug design and development, material science, and biophysics. By providing a neural framework to study the motion and interactions of particles in Brownian systems, the proposed work offers a versatile tool for scientific exploration, innovation, and problem-solving in various fields, and ultimately driving advancements and discoveries with broad-ranging impacts.

\textbf{Limitations and future works:} In this work, we have used a message-passing graph convolutional network for modeling Brownian dynamics with the injection of appropriate inductive biases. How does the performance change with the injection of attention on edges? Is it important to enforce equivariance to rotation through the usage of an equivariant graph? Can we make the neural network interpretable by uncovering the functions it is learning? These are interesting open questions that can be pursued as future work. 

Further, from a physics point of view, the work focuses on Brownian dynamics, which is an over-damped limit of the Langevin equation. How does the model perform in the case of a Langevin equation where the  acceleration is non-zero? How can the model be extended to other SDEs? We aim to explore these questions in our future work. Further, the present work does not analyze the effect of different integrators. This can also be pursued as part of future work.

\newpage
\bibliographystyle{plain}
\bibliography{references}

\begin{thebibliography}{10}

\bibitem{rigidbody}
Ravinder Bhattoo, Sayan Ranu, and NM~Krishnan.
\newblock Learning articulated rigid body dynamics with lagrangian graph neural
  network.
\newblock {\em Advances in Neural Information Processing Systems},
  35:29789--29800, 2022.

\bibitem{lgnn}
Ravinder Bhattoo, Sayan Ranu, and NM~Anoop Krishnan.
\newblock Learning the dynamics of particle-based systems with lagrangian graph
  neural networks.
\newblock {\em Machine Learning: Science and Technology}, 2023.

\bibitem{gnode}
Suresh Bishnoi, Ravinder Bhattoo, Sayan Ranu, and NM~Krishnan.
\newblock Enhancing the inductive biases of graph neural ode for modeling
  dynamical systems.
\newblock {\em International Conference on Learning Representations}, 2022.

\bibitem{bradbury2020jax}
James Bradbury, Roy Frostig, Peter Hawkins, Matthew~James Johnson, Chris Leary,
  Dougal Maclaurin, and Skye Wanderman-Milne.
\newblock Jax: composable transformations of python+ numpy programs, 2018.
\newblock {\em URL http://github. com/google/jax}, 4:16, 2020.

\bibitem{branka1998algorithms}
AC~Bra{\'n}ka and DM~Heyes.
\newblock Algorithms for brownian dynamics simulation.
\newblock {\em Physical Review E}, 58(2):2611, 1998.

\bibitem{cates1985brownian}
ME~Cates.
\newblock Brownian dynamics of self-similar macromolecules.
\newblock {\em Journal de Physique}, 46(7):1059--1077, 1985.

\bibitem{chen2018neural}
Ricky~TQ Chen, Yulia Rubanova, Jesse Bettencourt, and David Duvenaud.
\newblock Neural ordinary differential equations.
\newblock In {\em Proceedings of the 32nd International Conference on Neural
  Information Processing Systems}, pages 6572--6583, 2018.

\bibitem{chung1999permeation}
Shin-Ho Chung, Toby~W Allen, Matthew Hoyles, and Serdar Kuyucak.
\newblock Permeation of ions across the potassium channel: Brownian dynamics
  studies.
\newblock {\em Biophysical Journal}, 77(5):2517--2533, 1999.

\bibitem{langevin}
W.~T. Coffey and Yu~P. Kalmykov.
\newblock {\em The Langevin Equation: With Applications to Stochastic Problems
  in Physics, Chemistry and Electrical Engineering (Third edition)}, volume~27
  of {\em World Scientific Series in Contemporary Chemical Physics}.
\newblock World Scientific, Trinity College, Dublin, Ireland and Université de
  Perpignan, France, 2020.

\bibitem{cranmer2020discovering}
Miles Cranmer, Alvaro Sanchez~Gonzalez, Peter Battaglia, Rui Xu, Kyle Cranmer,
  David Spergel, and Shirley Ho.
\newblock Discovering symbolic models from deep learning with inductive biases.
\newblock {\em Advances in Neural Information Processing Systems}, 33, 2020.

\bibitem{lnn1}
Marc Finzi, Ke~Alexander Wang, and Andrew~G Wilson.
\newblock Simplifying hamiltonian and lagrangian neural networks via explicit
  constraints.
\newblock In H.~Larochelle, M.~Ranzato, R.~Hadsell, M.~F. Balcan, and H.~Lin,
  editors, {\em Advances in Neural Information Processing Systems}, volume~33,
  pages 13880--13889. Curran Associates, Inc., 2020.

\bibitem{gabdoulline2001protein}
Razif~R Gabdoulline and Rebecca~C Wade.
\newblock Protein-protein association: investigation of factors influencing
  association rates by brownian dynamics simulations.
\newblock {\em Journal of molecular biology}, 306(5):1139--1155, 2001.

\bibitem{genkin2021learning}
Mikhail Genkin, Owen Hughes, and Tatiana~A Engel.
\newblock Learning non-stationary langevin dynamics from stochastic
  observations of latent trajectories.
\newblock {\em Nature Communications}, 12(1):5986, 2021.

\bibitem{gnesotto2020learning}
Federico~S Gnesotto, Grzegorz Gradziuk, Pierre Ronceray, and Chase~P Broedersz.
\newblock Learning the non-equilibrium dynamics of brownian movies.
\newblock {\em Nature communications}, 11(1):5378, 2020.

\bibitem{jraph2020github}
Jonathan Godwin*, Thomas Keck*, Peter Battaglia, Victor Bapst, Thomas Kipf,
  Yujia Li, Kimberly Stachenfeld, Petar Veli\v{c}kovi\'{c}, and Alvaro
  Sanchez-Gonzalez.
\newblock {J}raph: {A} library for graph neural networks in jax., 2020.

\bibitem{greydanus2019hamiltonian}
Samuel Greydanus, Misko Dzamba, and Jason Yosinski.
\newblock Hamiltonian neural networks.
\newblock {\em Advances in Neural Information Processing Systems},
  32:15379--15389, 2019.

\bibitem{gruver2021deconstructing}
Nate Gruver, Marc~Anton Finzi, Samuel~Don Stanton, and Andrew~Gordon Wilson.
\newblock Deconstructing the inductive biases of hamiltonian neural networks.
\newblock In {\em International Conference on Learning Representations}, 2021.

\bibitem{helfand1980brownian}
Eugene Helfand, ZR~Wasserman, and Thomas~A Weber.
\newblock Brownian dynamics study of polymer conformational transitions$\pm$.
\newblock {\em Macromolecules}, 13(3):526--533, 1980.

\bibitem{huang2021equivariant}
Wenbing Huang, Jiaqi Han, Yu~Rong, Tingyang Xu, Fuchun Sun, and Junzhou Huang.
\newblock Equivariant graph mechanics networks with constraints.
\newblock In {\em International Conference on Learning Representations}, 2022.

\bibitem{karniadakis2021physics}
George~Em Karniadakis, Ioannis~G Kevrekidis, Lu~Lu, Paris Perdikaris, Sifan
  Wang, and Liu Yang.
\newblock Physics-informed machine learning.
\newblock {\em Nature Reviews Physics}, 3(6):422--440, 2021.

\bibitem{karplus1994protein}
Martin Karplus and David~L Weaver.
\newblock Protein folding dynamics: The diffusion-collision model and
  experimental data.
\newblock {\em Protein science}, 3(4):650--668, 1994.

\bibitem{klank2018brownian}
Rebecca~L Klank, Steven~S Rosenfeld, and David~J Odde.
\newblock A brownian dynamics tumor progression simulator with application to
  glioblastoma.
\newblock {\em Convergent science physical oncology}, 4(1):015001, 2018.

\bibitem{metahamiltonian}
Seungjun Lee, Haesang Yang, and Woojae Seong.
\newblock Identifying physical law of hamiltonian systems via meta-learning.
\newblock In {\em International Conference on Learning Representations}, 2021.

\bibitem{madura1995electrostatics}
Jeffry~D Madura, James~M Briggs, Rebecca~C Wade, Malcolm~E Davis, Brock~A Luty,
  Andrew Ilin, Jan Antosiewicz, Michael~K Gilson, Babak Bagheri, L~Ridgway
  Scott, et~al.
\newblock Electrostatics and diffusion of molecules in solution: simulations
  with the university of houston brownian dynamics program.
\newblock {\em Computer Physics Communications}, 91(1-3):57--95, 1995.

\bibitem{ni2022ntfields}
Ruiqi Ni and Ahmed~H Qureshi.
\newblock Ntfields: Neural time fields for physics-informed robot motion
  planning.
\newblock {\em arXiv preprint arXiv:2210.00120}, 2022.

\bibitem{noguchi2001self}
Hiroshi Noguchi and Masako Takasu.
\newblock Self-assembly of amphiphiles into vesicles: a brownian dynamics
  simulation.
\newblock {\em Physical Review E}, 64(4):041913, 2001.

\bibitem{northrup1992kinetics}
Scott~H Northrup and Harold~P Erickson.
\newblock Kinetics of protein-protein association explained by brownian
  dynamics computer simulation.
\newblock {\em Proceedings of the National Academy of Sciences},
  89(8):3338--3342, 1992.

\bibitem{park2021accurate}
Cheol~Woo Park, Mordechai Kornbluth, Jonathan Vandermause, Chris Wolverton,
  Boris Kozinsky, and Jonathan~P Mailoa.
\newblock Accurate and scalable graph neural network force field and molecular
  dynamics with direct force architecture.
\newblock {\em npj Computational Materials}, 7(1):1--9, 2021.

\bibitem{sanchez2019hamiltonian}
Alvaro Sanchez-Gonzalez, Victor Bapst, Kyle Cranmer, and Peter Battaglia.
\newblock Hamiltonian graph networks with ode integrators.
\newblock {\em arXiv e-prints}, pages arXiv--1909, 2019.

\bibitem{sanchez2020learning}
Alvaro Sanchez-Gonzalez, Jonathan Godwin, Tobias Pfaff, Rex Ying, Jure
  Leskovec, and Peter Battaglia.
\newblock Learning to simulate complex physics with graph networks.
\newblock In {\em International Conference on Machine Learning}, pages
  8459--8468. PMLR, 2020.

\bibitem{schoenholz2020jax}
Samuel Schoenholz and Ekin~Dogus Cubuk.
\newblock Jax md: a framework for differentiable physics.
\newblock {\em Advances in Neural Information Processing Systems}, 33, 2020.

\bibitem{benchmarking}
Abishek Thangamuthu, Gunjan Kumar, Suresh Bishnoi, Ravinder Bhattoo,
  NM~Krishnan, and Sayan Ranu.
\newblock Unravelling the performance of physics-informed graph neural networks
  for dynamical systems.
\newblock {\em Advances in Neural Information Processing Systems}, 2022.

\bibitem{vamathevan2019applications}
Jessica Vamathevan, Dominic Clark, Paul Czodrowski, Ian Dunham, Edgardo Ferran,
  George Lee, Bin Li, Anant Madabhushi, Parantu Shah, Michaela Spitzer, et~al.
\newblock Applications of machine learning in drug discovery and development.
\newblock {\em Nature reviews Drug discovery}, 18(6):463--477, 2019.

\bibitem{van1982algorithms}
WF~Van~Gunsteren and HJC Berendsen.
\newblock Algorithms for brownian dynamics.
\newblock {\em Molecular Physics}, 45(3):637--647, 1982.

\bibitem{zhong2019symplectic}
Yaofeng~Desmond Zhong, Biswadip Dey, and Amit Chakraborty.
\newblock Symplectic ode-net: Learning hamiltonian dynamics with control.
\newblock In {\em International Conference on Learning Representations}, 2019.

\bibitem{zhong2021benchmarking}
Yaofeng~Desmond Zhong, Biswadip Dey, and Amit Chakraborty.
\newblock Benchmarking energy-conserving neural networks for learning dynamics
  from data.
\newblock In {\em Learning for Dynamics and Control}, pages 1218--1229. PMLR,
  2021.

\end{thebibliography}
\clearpage
\section*{Appendix}
\label{sec:paper_appendix}

\appendix
\renewcommand{\thesubsection}{\Alph{subsection}}
\renewcommand{\thefigure}{\Alph{figure}}
\renewcommand{\thetable}{\Alph{table}}

\subsection{Proof of Thm.~\ref{thm:mom_cons}}
\label{app:proof}
\textsc{Proof.} Consider a system without any external field where the dynamics is governed only by the internal deterministic forces, for instance, a $n$-spring system and stochastic force terms representing the interaction with the environment, for example, spring-particle systems in fluid system with the stochastic term representing the collision of the particles with the atoms of the fluid environment. Note that the deterministic term consisting of the forces is also termed as the drift term as it contributed to the overall drift of the system, if any. Accordingly, the stochastic term is referred to as the diffusion term.
Since, the acceleration of each of the particles is zero, the net force on the system should be zero, that is, the total linear momentum of the system due to the internal deterministic forces remain conserved. The conservation of linear momentum for this system implies that $\sum_{i=1}^n \hat{F}_i = 0 $. 
Assume that the total error between the predicted and actual forces of all the particles at time $t$ to be $\hat \varepsilon_{t} $. To prove that \name{} conserve the linear momentum exactly, it is sufficient to prove that $\hat \varepsilon_{t} = 0$, since the stochastic term has zero mean. Consider,
\begin{align}
\label{eq:acc}
    \hat \varepsilon_{t} &= \sum_{i=1}^n \hat \varepsilon_{i,t} = \sum_{i=1}^n(\hat{F}_i - F_i) =  \sum_{i=1}^n\hat{F}_i - \sum_{i=1}^nF_i
\end{align}
Note that in Brownian dynamics, which is the over damped limit of Langevin equation (Eq.~\ref{eq:langevin}), the acceleration of each particle is assumed to be zero. Thus, without loss of generality, $\sum_{i=1}^nF_i = F_i = 0$, in the absence of an external field. Hence,
\begin{align*}
    \hat \varepsilon_{t} &= \sum_{i=1}^n \hat \varepsilon_{i,t} = \sum_{i=1}^n\hat{F}_i = \sum_{i=1}^n \left( \sum_{j \in \mathcal{N}^{in}_i} \hat{F}_{ji} + \sum_{j \in \mathcal{N}^{out}_i} -\hat{F}_{ij} \right) = 0 \hspace{1in}\square 
\end{align*}
Thus, the sum of the predicted force of the system remains zero leading to the conservation of linear momentum of the system. 
\subsection{Euler Maruyama Integrator}
\label{app:em_int}
Consider a system of $N$ particles having masses $M$ and interacting with each other through an interaction potential $U(X(t))$, where $X(t)$ represents the position vectors of all the particles at time $t$, that constitute a time-dependent random variable. The Taylor series expansion of the position of a particle $i$ at time $t + \Delta t$ is given by
\begin{equation}
    X_{i}(t + \Delta t) = X_{i}(t) + \Delta t \dot{X}_{i}(t) + \Delta t^2 \ddot{X}_{i}(t) + \ldots 
\label{eq:taylor}
\end{equation}
In the over-damped limit of Langevin dynamics, namely, Brownian dynamics, acceleration and higher order terms are assumed to be zero. Thus, substituting for velocity given by Eq.~\ref{eq:langevinsimple} in Eq.~\ref{eq:taylor}, we get
\begin{equation}
 X_i(t+\Delta t) = X_i(t) + F_i/\gamma_i \Delta t + \sqrt{\frac{2k_BT}{\gamma_i}} \Delta \Omega_t
 \label{eq:Euler_Maruyama}
 \end{equation}
where $\Delta \Omega_t$ is a random number sampled from a standard Normal distribution. Note that the Euler-Maruyama is the simplest form of a stochastic numerical integrator. Further, higher order integrals can also be employed to integrate the SDEs.

\subsection{Experimental systems}
\label{app:systems}
\subsubsection{Simulation Environment}
 All the simulations and training were carried out in the JAX environment~\cite{schoenholz2020jax,bradbury2020jax}. The graph architecture was developed using the jraph package~\cite{jraph2020github}. All the codes used in the present work will be made available upon the acceptance of the work.

\subsubsection{Linear \textit{n}-spring system}
In this system, $n$-point masses are connected by elastic springs that deform linearly with extension or compression. Note that similar to a pendulum setup, each mass $m_i$ is connected only to two masses $m_{i-1}$ and $m_{i+1}$ through springs so that all the masses form a closed connection. Thus, the deterministic force experienced by each mass $i$ due to a spring connecting $i$ to its neighbor $j$ is $F_{ij}=-k(||X_{i}-X_{j}||-R_{ij})$, where $R_{ij}$ is the equilibrium length of the spring and $k$ represent the un-deformed length and the stiffness, respectively, of the spring.

\subsubsection{Non-linear \textit{n}-spring system}
This system is similar to a linear $n$-spring system with the difference that the spring force is non-linear. Specifically, the deterministic force experienced by each mass $i$ due to a spring connecting $i$ to its neighbor $j$ is $F_{ij}=-k(||X_{i}-X_{j}||-R_{ij})^3$, where $R_{ij}$ is the equilibrium length of the spring and $k$ represent the un-deformed length and the stiffness, respectively, of the spring.

\subsubsection{Binary linear \textit{n}-spring system}
This system is similar to a linear $n$-spring system with the difference that the system consists of two different types of particles with different masses, friction coefficient, and $\gamma$. Thus, the deterministic force experienced by each mass $i$ due to a spring connecting $i$ to its neighbor $j$ remains the same as $F_{ij}=-k(||X_{i}-X_{j}||-R_{ij})$, where $R_{ij}$ is the equilibrium length of the spring and $k$ represent the un-deformed length and the stiffness, respectively, of the spring. However, the stochastic part varies depending on the type of the particle.

\subsection{Dataset generation}
\label{app:imple}
All the datasets are generated using the known deterministic forces of the systems, along with the stochastic, as described in Section~\ref{app:systems} and Eq.\ref{eq:langevinsimple}. For each system, we create the training data by performing forward simulations with 100 random initial conditions. For both linear and nonlinear $n-$spring system, $n = 5$ is used for generating the training data. For the binary system, a 10 particle system is used for generating the data with the ratio of particle type 1 to type 2 as 3:7. A timestep of $10^{-3}s$ is used to integrate the equations of motion for all the systems. The Euler-Maruyama algorithm is used to integrate equations of motion due to its ability to handle stochastic differential equations. The details of the parameters used for each of the systems are given below.\\
$\bullet$ \textbf{Linear $n$-spring system}
\begin{center}
\begin{tabular}{ |c|c| } 
 \hline
 \textbf{Parameter} & \textbf{Value} \\ 
 \hline
 Mass ($M$) & 1 unit \\ 
 Stiffness ($k$) & 1 unit \\ 
 Damping constant ($\zeta$) & 1 unit\\
 $k_BT$ & 1 unit \\
 Equilibrium length of the spring ($R_{ij}$) & 1 unit\\
 \hline
\end{tabular}
\end{center}

$\bullet$ \textbf{Nonlinear $n$-spring system}
\begin{center}
\begin{tabular}{ |c|c| } 
 \hline
 \textbf{Parameter} & \textbf{Value} \\ 
 \hline
 Mass ($M$) & 1 unit \\ 
 Stiffness ($k$) & 1 unit \\ 
 Damping constant ($\zeta$) & 1 unit\\
 $k_BT$ & 1 unit \\
 Equilibrium length of the spring ($R_{ij}$) & 1 unit\\
 \hline
\end{tabular}
\end{center}

$\bullet$ \textbf{Binary linear $n$-spring system}
\begin{center}
\begin{tabular}{ |c|c| } 
 \hline
 \textbf{Parameter} & \textbf{Value} \\ 
 \hline
 Mass ($M$) & 1 unit \\ 
 Stiffness ($k$) & 1 unit \\ 
 ($\gamma_1$) & 1 unit\\
 ($\gamma_2$) & 2 unit\\
 $k_BT$ & 1 unit \\
 Equilibrium length of the spring ($R_{ij}$) & 1 unit\\
 \hline
\end{tabular}
\end{center}

From the 100 simulations for the systems obtained from the rollout starting from 100 random initial conditions, 100 data points are extracted per simulation, resulting in a total of 10000 data points. For training, we create several sets of trajectories with positions from two consecutive timesteps $t$ and $t + \Delta t$, where the input for the model is the positions of all the particles at $t$ and the output against which the model is trained is the position of all the particles at $t + \Delta t$ employing the loss function as defined in Eq.~\ref{eq:lossfunction}. Note that in the present work, we use only trajectories of length one timestep only. The trained models are evaluated on 1000 trajectories generated from 100 different initial conditions with each initial condition simulated for 10 random seeds, all of which are unseen during the training. The trajectories considered are of 100 timesteps, that is, 0.1 s. Note that this is not a task the models are trained for; the training was done only on one timestep trajectories.  All the error metrics presented in the work are averaged over the 1000 trajectories for each of the models and casees unless specified otherwise.

For zero-shot generalizability, for each case of different system sizes and temperatures, we directly evaluate the trained models on 1000 trajectories. No fine tuning or additional training is performed for any of the different system sizes or temperatures.

\subsection{Training details and hyper-parameters}
\label{app:hyper}
The detailed procedures followed for the training of the models and the hyperparameters employed for each of the models, identified based on good practices, are provided in this section. 

\subsubsection{Training details}
The training dataset is divided in 80:20 ratio randomly, where the 80\% is used for training and 20\% is used as the validation set for hyperparametric optimization. Further, the trained models are tested on its ability to predict statistically equivalent trajectory, a task it was not trained on. Specifically, the systems are evaluated on a $0.1 s$ long trajectory, that is $100\times$ larger than the training trajectory. The error metrics are computed based on 1000 different trajectories from 100 initial conditions, each run with 10 random seeds. All models are trained till the losses saturated. A learning rate of $10^{-3}$ was used with the Adam optimizer for the training. Detailed hyperparameters employed for all the models are provided next.\\

\subsubsection{Hyper-parameters}
The hyper-parameters used for different models, namely, NN, \bnn, \bfgn, \bdgnn, and \name are given in the tables below.\\

$\bullet$\textbf{NN, \bnn}
\begin{center}
\begin{tabular}{ |c|c| } 
 \hline
 \textbf{Parameter} & \textbf{Value} \\ 
 \hline
 Hidden layer neurons (MLP) & 16 \\ 
 Number of hidden layers (MLP) & 2 \\ 
 Activation function & squareplus\\
 Optimizer & ADAM \\
 Learning rate & $1.0e^{-3}$ \\
 Batch size & 20 \\
 \hline
\end{tabular}
\end{center}

$\bullet$\textbf{\bfgn}
\begin{center}
\begin{tabular}{ |c|c| } 
 \hline
 \textbf{Parameter} & \textbf{Value} \\
 \hline
 Node embedding dimension & 8 \\ 
 Edge embedding dimension & 8 \\ 
 Hidden layer neurons (MLP) & 16 \\ 
 Number of hidden layers (MLP) & 2 \\ 
 Activation function & squareplus\\
 Number of layers of message passing & 1\\
 Optimizer & ADAM \\
 Learning rate & $1.0e^{-3}$ \\
 Batch size & 20 \\
 \hline
\end{tabular}
\end{center}

$\bullet$\textbf{\bdgnn, \name}
\begin{center}
\begin{tabular}{ |c|c| } 
 \hline
 \textbf{Parameter} & \textbf{Value} \\ 
 \hline
 Node embedding dimension & 5 \\ 
 Edge embedding dimension & 5 \\ 
 Hidden layer neurons (MLP) & 5 \\ 
 Number of hidden layers (MLP) & 2 \\ 
 Activation function & squareplus\\
 Number of layers of message passing & 1\\
 Optimizer & ADAM \\
 Learning rate & $1.0e^{-3}$ \\
 Batch size & 20 \\
 \hline
\end{tabular}
\end{center}

\subsection{Baselines}
\label{app:baselines}

\begin{figure}
\centering
\includegraphics[width=\columnwidth]{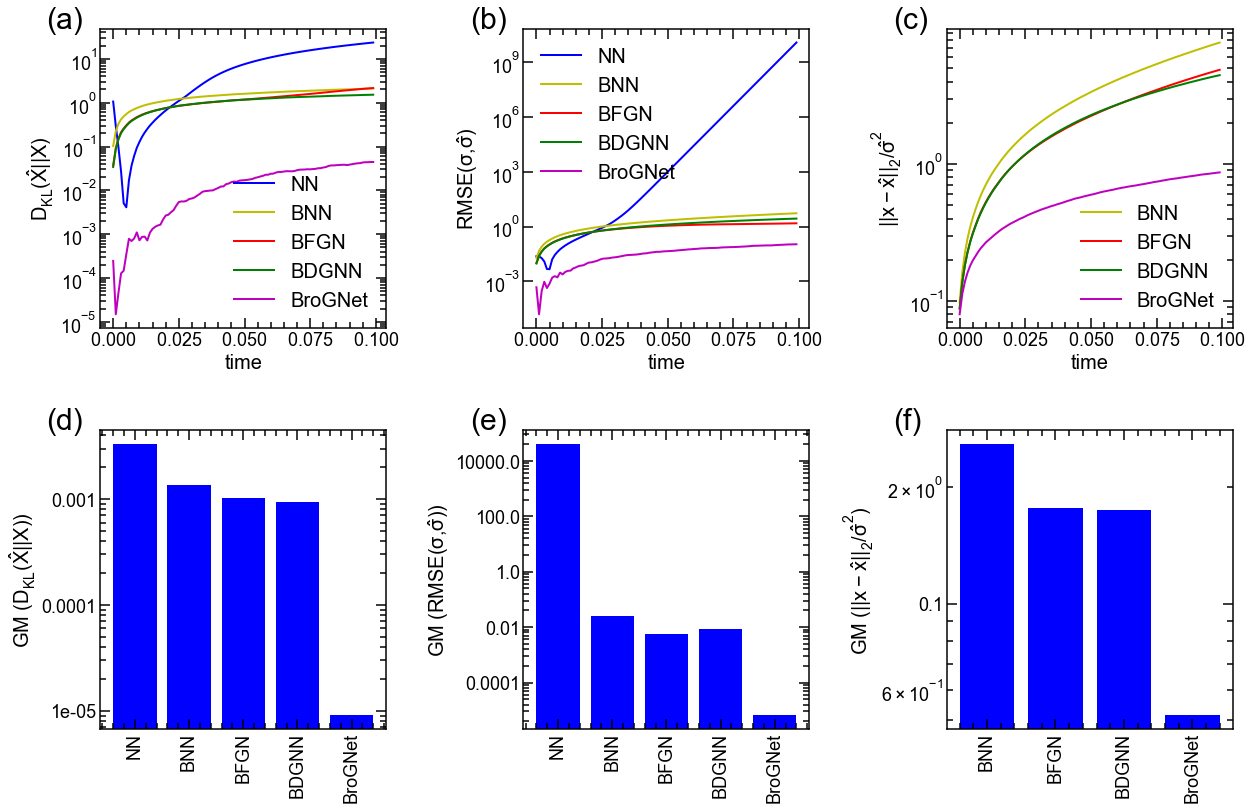}
\caption{(a) Trajectory error, (b) Brownian error, (c) position error and Geometric mean of (d) trajectory error, (e) Brownian error, and (f) position error of NN, \bnn, \bfgn, \bdgnn, and \name trained on 10000 data points for linear spring. All the results are evaluated based on 1000 forward simulations from 100 initial conditions, each evaluated with 10 random seeds.}
\label{fig:fig_baselines_a}
\end{figure}

\begin{figure}
\centering
\includegraphics[width=\columnwidth]{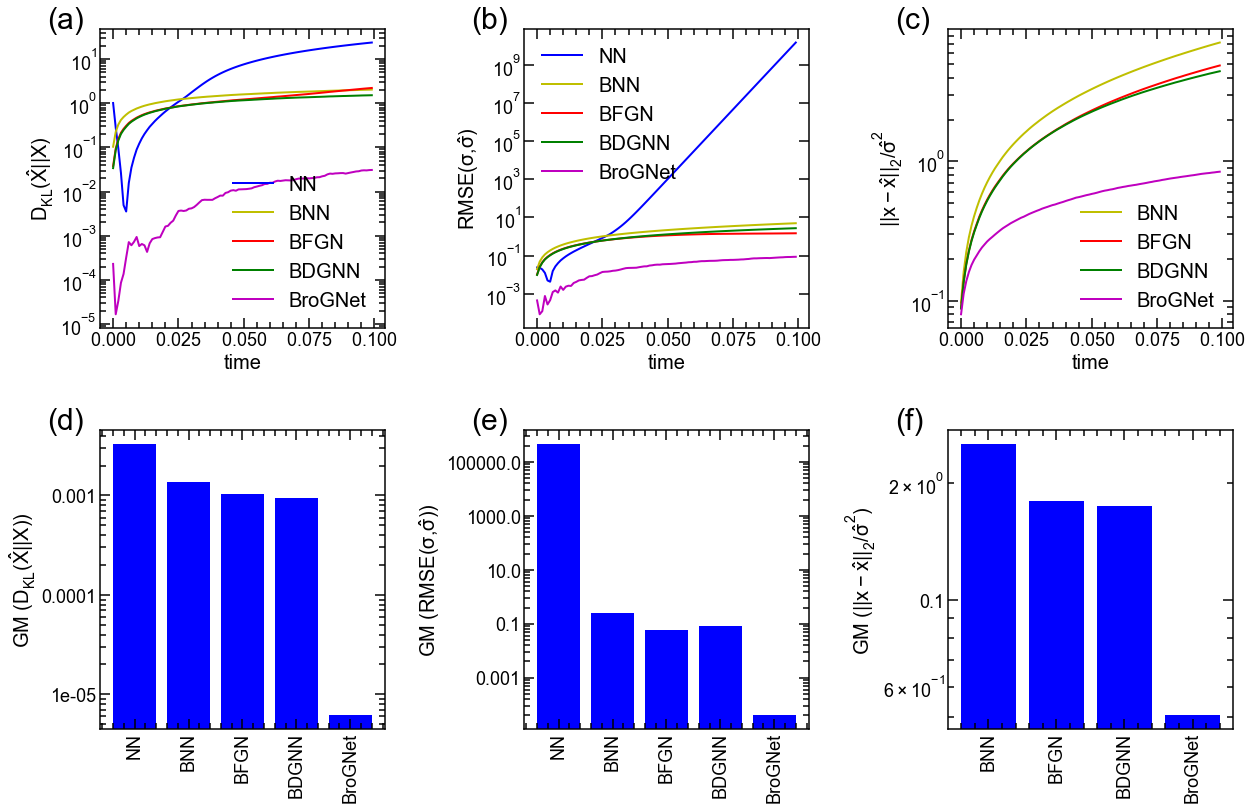}
\caption{(a) Trajectory error, (b) Brownian error, (c) position error and Geometric mean of (d) trajectory error, (e) Brownian error, and (f) position error of NN, \bnn, \bfgn, \bdgnn, and \name trained on 10000 data points for non-linear spring. All the results are evaluated based on 1000 forward simulations from 100 initial conditions, each evaluated with 10 random seeds.}
\label{fig:fig_baselines_b}
\end{figure}

\begin{figure}
\centering
\includegraphics[width=\columnwidth]{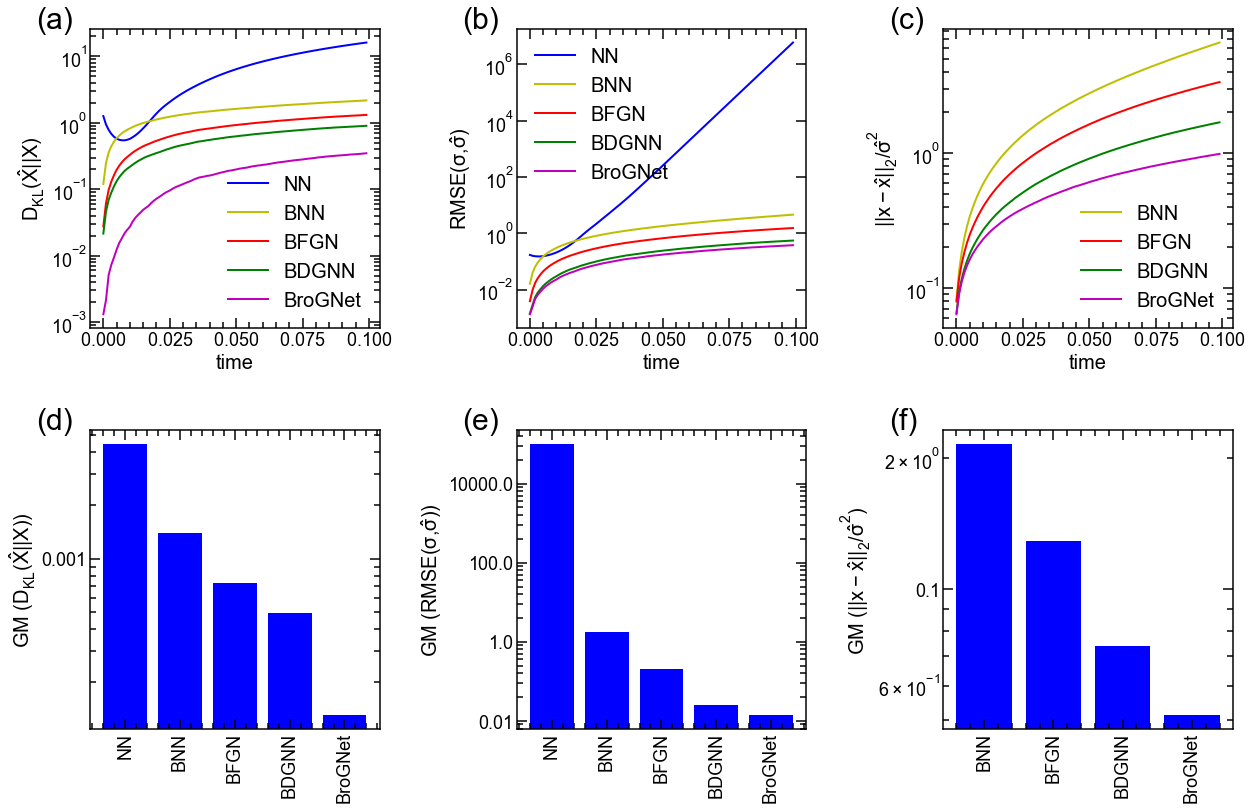}
\caption{(a) Trajectory error, (b) Brownian error, (c) position error and Geometric mean of (d) trajectory error, (e) Brownian error, and (f) position error of NN, \bnn, \bfgn, \bdgnn, and \name trained on 10000 data points for binary spring. All the results are evaluated based on 1000 forward simulations from 100 initial conditions, each evaluated with 10 random seeds.}
\label{fig:fig_baselines_c}
\end{figure}

Figures~\ref{fig:fig_baselines_a}~\ref{fig:fig_baselines_b} and \ref{fig:fig_baselines_c} show the trajectory error, Brownian error, and the position error and the geometric mean of these errors over the trajectory for all the baselines, namely, NN, \bnn, \bfgn, and \bdgnn, in comparison to \name for linear $5-$spring system, non-linear $5-$spring system, and binary linear $10-$spring systems, respectively. We observe that \name significantly outperforms other models.


\subsection{Zero-shot generalizability}
\label{app:errors}

\begin{figure}
\centering
\includegraphics[width=\columnwidth]{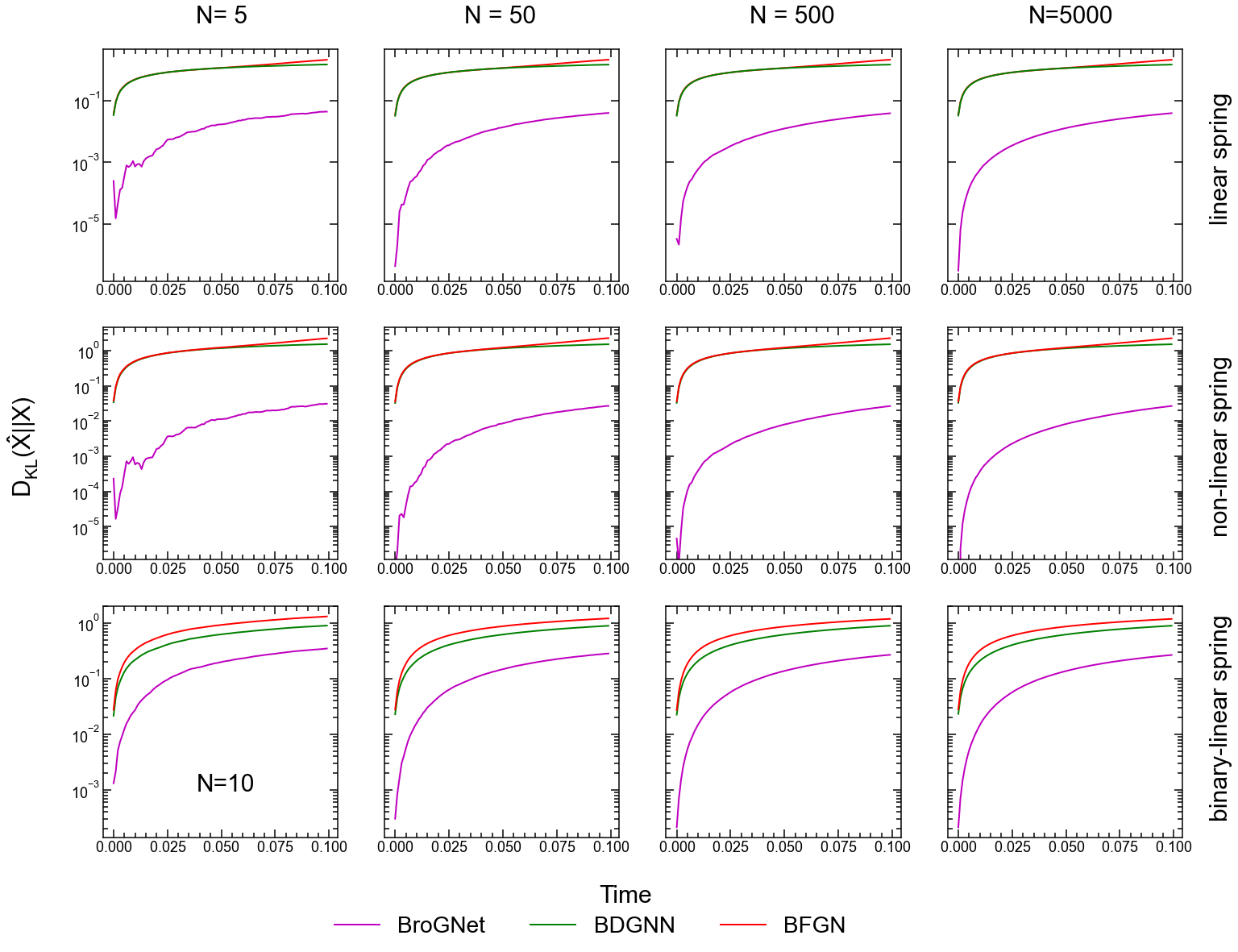}
\caption{Rollout error of \name, \bdgnn, and \bfgn tested on N = 50, 500, 5000 for linear (row 1), non-linear spring (row 2), and binary linear spring (row 3) systems evaluated on 1000 forward trajectories. Note that for linear and non-linear systems, the models are trained on 5-particle systems whereas for binary system, the models are trained on 10-particle system.}
\label{fig:fig_zeroshot_1}
\end{figure}

\begin{figure}
\centering
\includegraphics[width=\columnwidth]{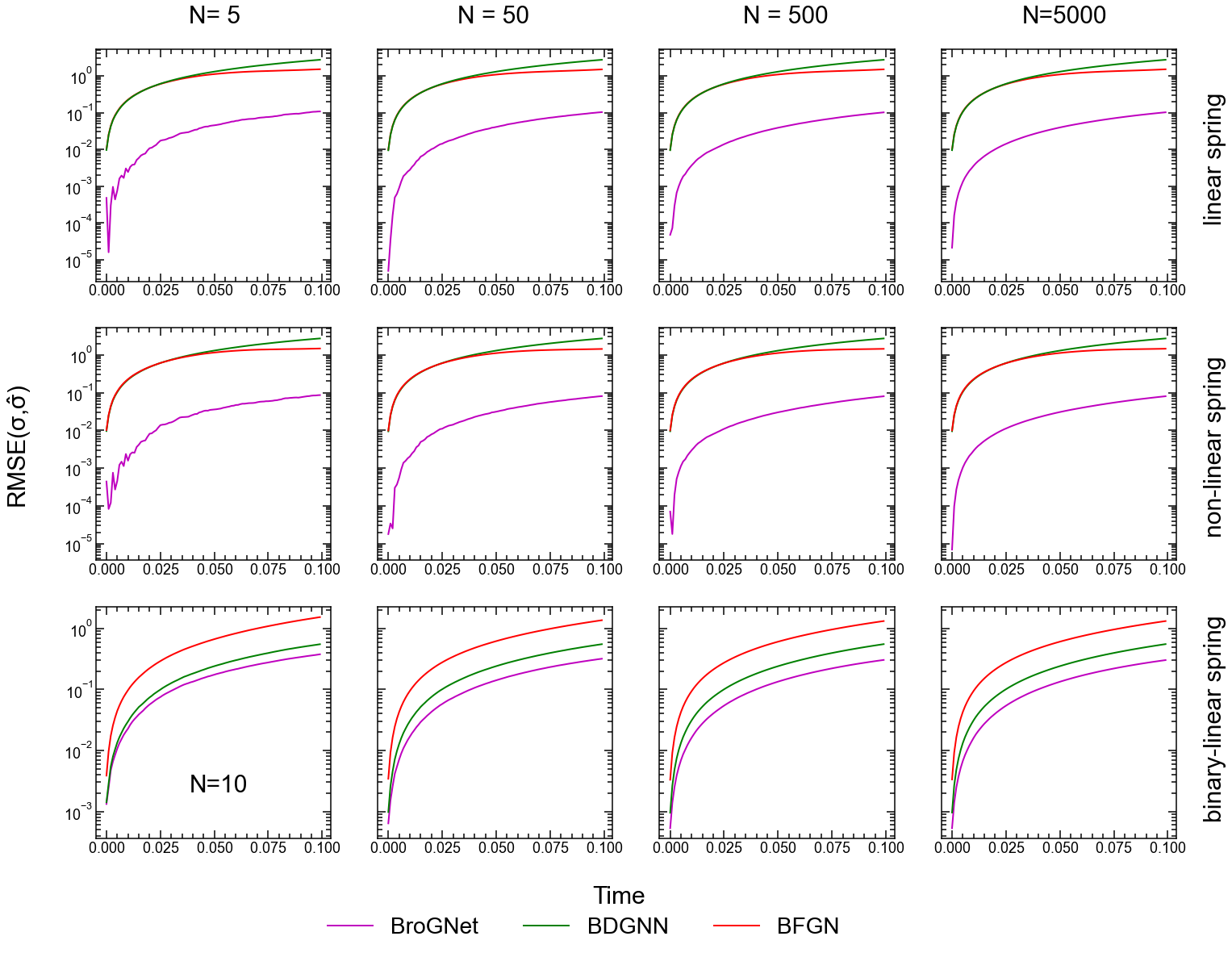}
\caption{Brownian error of \name, \bdgnn, and \bfgn trained on N = 5 system and tested on N = 50, 500, 5000 for linear (row 1), non-linear spring (row 2), and binary linear spring (row 3) systems evaluated on 1000 forward trajectories.}
\label{fig:fig_zeroshot_2}
\end{figure}

\begin{figure}
\centering
\includegraphics[width=\columnwidth]{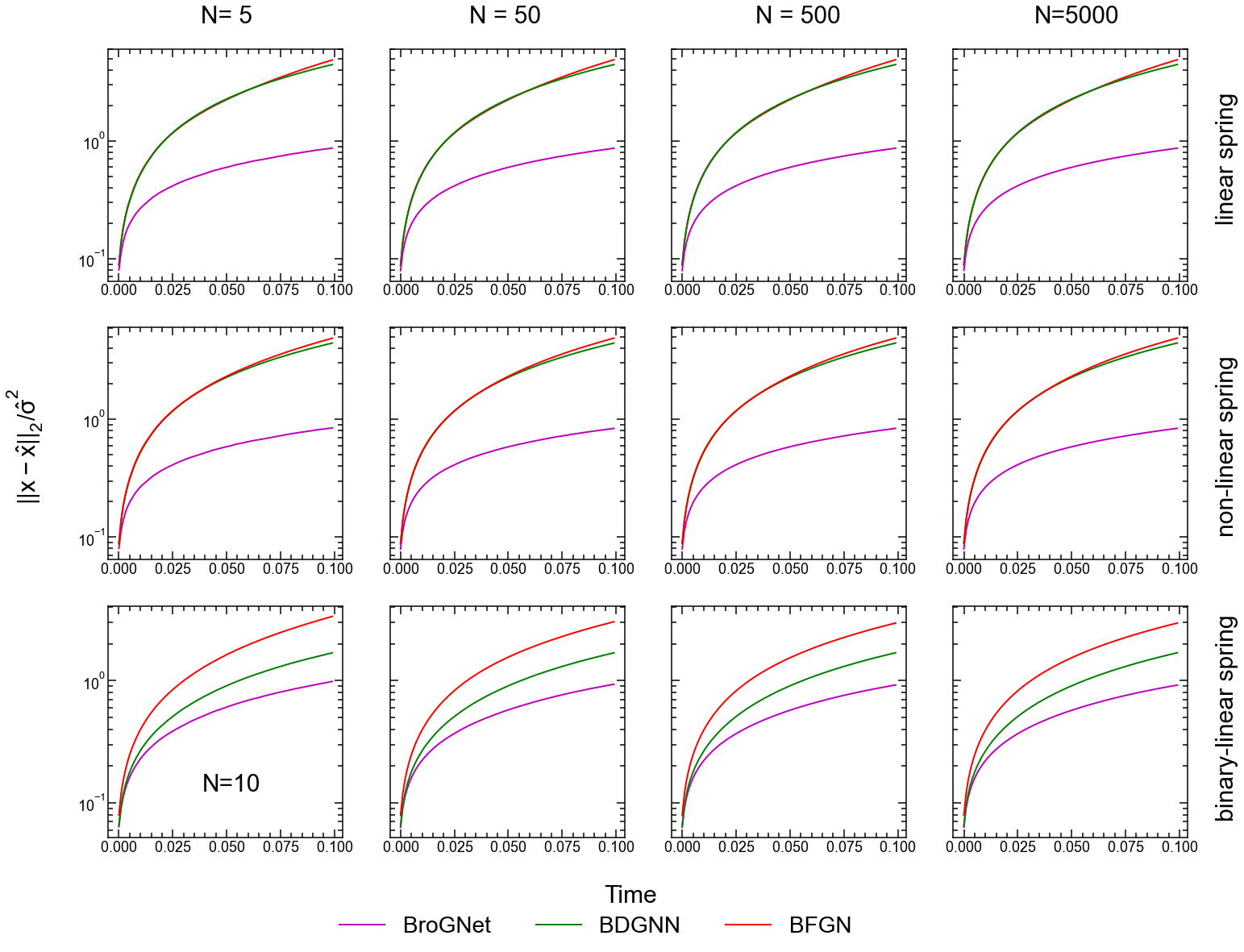}
\caption{Position error of \name, \bdgnn, and \bfgn trained on N = 5 system and tested on N = 50, 500, 5000 for linear (row 1), non-linear spring (row 2), and binary linear spring (row 3) systems evaluated on 1000 forward trajectories.}
\label{fig:fig_zeroshot_3}
\end{figure}

Here, we evaluate the zero-shot generalizability of \name in comparison to \bdgnn and \bfgn for linear, non-linear, and binary linear $n-$ spring systems. Figures~ \ref{fig:fig_zeroshot_1}~\ref{fig:fig_zeroshot_2} and~\ref{fig:fig_zeroshot_3} show the trajectory, Brownian, and position error, respectively, of \bdgnn, \bfgn, and \name evaluated on 50, 500, and 5000 particle systems. Note for linear and non-linear springs, the models are trained on 5-particle systems, while for binary linear spring system, the models are trained on 10 particle systems. We observe that \name significantly outperforms other models.

\subsection{Training and data efficiency}
\label{app:data_eff}

\begin{figure}
\centering
\includegraphics[width=\columnwidth]{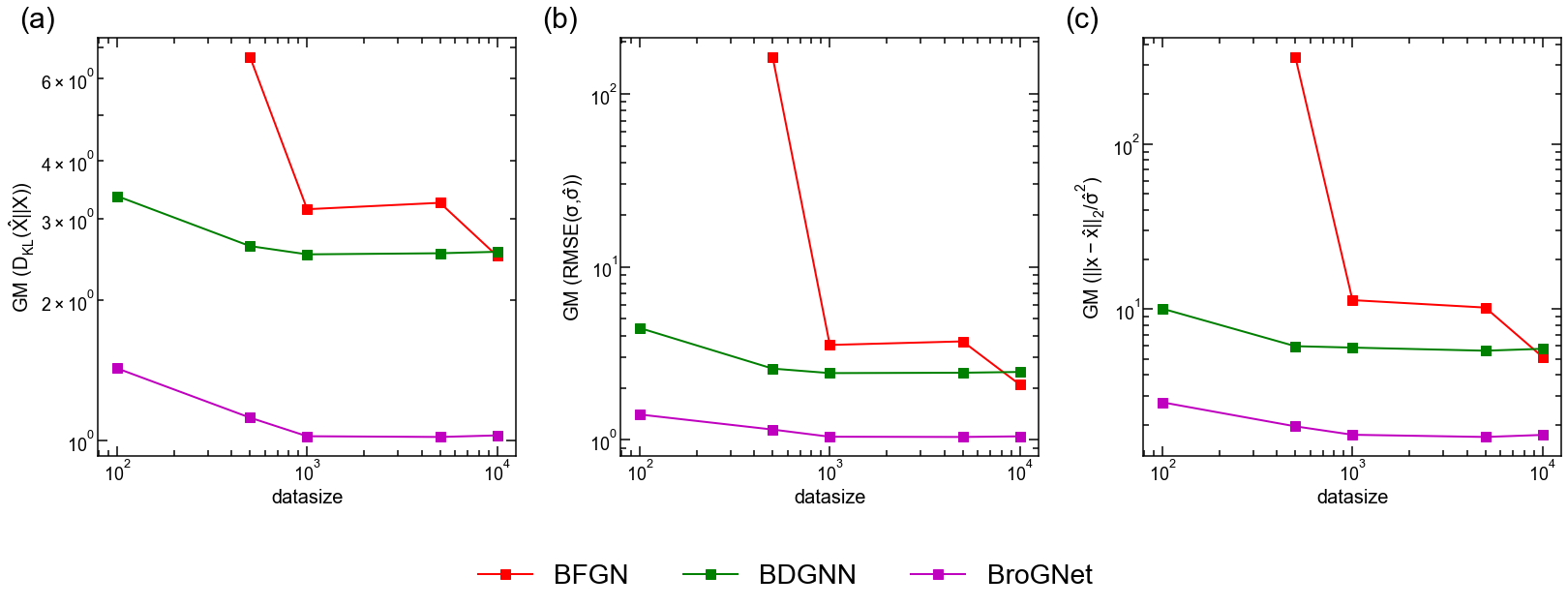}
\caption{Geometric mean of (a) trajectory error, (b) Brownian error, and (c) position error of \bfgn, \bdgnn, and \name trained on varying dataset size. All the results are evaluated based on 1000 forward simulations based on 100 initial conditions, each evaluated with 10 random seeds.}
\label{fig:fig_GM}
\end{figure}

We also evaluated the computational efficiency and data efficiency of \name in comparison to the baselines. 
Specifically, we evaluated the performance of \name with dataset size varying as 100, 500, 1000, 5000, and 10000 in comparison to the baselines \bfgn, and \bdgnn. The results shown in Fig.~\ref{fig:fig_GM} suggest that \name is indeed able to learn efficiently from a small dataset size in comparison to the baselines. Further, we also evaluate the training time and forward simulation time of \name for different system sizes. The results suggest that \name is scalable to large system sizes.

\begin{table}[h]
\centering
\begin{tabular}{ |l|c|c| }
\toprule
\textbf{Models} & \textbf{Training time (in sec)} & \textbf{Forward Simulation time (in sec)} \\
\midrule

NN & 391 & 0.012\\
\bnn & 635 & 0.222\\
\bfgn & 1755 & 0.642\\
\bdgnn & 1934 & 0.411\\
\name & 2126 & 0.506\\
\bottomrule
\end{tabular}
\caption{Training time for 10,000 epochs and forward simulation time for 100 timesteps of NN, \bnn, \bfgn, \bdgnn, \name on linear 5 spring system. Note that the forward simulation time is computed as the average of 100 initial condition with each initial condition simulated for 10 different random seeds (altogether, an average of 1000 trajectories).}
\label{tab:springtime}
\end{table}

\end{document}